\documentclass[a4paper]{article}
\usepackage{arxiv}
\usepackage[utf8]{inputenc}
\usepackage[square,numbers]{natbib}
\usepackage[T1]{fontenc}
\usepackage[labelfont=bf,tableposition=top]{caption}
\usepackage{array,booktabs,tabularx,threeparttable}
\usepackage{float,placeins}
\usepackage{nicefrac}       
\usepackage{microtype} 
\usepackage{doi}
\usepackage{booktabs}       
\usepackage{amsfonts,amsmath,amssymb}       
\usepackage{xspace}
\usepackage{lipsum,amsmath,amssymb,amsfonts}
\usepackage{amsfonts}
\usepackage{tikz}
\usepackage{pgfplots}
\pgfplotsset{compat=1.18}
\usepackage{algorithm,algorithmic}
\usepackage{graphicx}
\usepackage{subcaption}
\usepackage{multirow}
\usepackage{booktabs}
\usepackage{multirow}
\usepackage{graphicx}
\graphicspath{ {./images/} }
\usepackage{hyperref}  
\title{BettiSplit: Topology-Guided Privacy-Aware Split Learning Against Feature Inversion and Gradient Leakage}

\author{
Akarsh K.~Nair\\
Department of Computer Science\\
Nottingham Trent University\\
Nottingham, United Kingdom\\
\texttt{akarsh.kongasserynair@ntu.ac.uk}
   \And
Muhammad Arifur Rahman\\
Department of Computer Science\\
Nottingham Trent University\\
Nottingham, United Kingdom\\
\texttt{arif.rahman@ntu.ac.uk}
  \And
David Brown\\
Department of Computer Science\\
Nottingham Trent University\\
Nottingham, United Kingdom\\
\texttt{david.brown@ntu.ac.uk}
   \And
Mufti Mahmud\\
Information and Computer Science Department\\
SDAIA--KFUPM Joint Research Center for AI\\
Interdisciplinary Research Center for Biosystems and Machines\\
King Fahd University of Petroleum \& Minerals\\
Dhahran 31261, Saudi Arabia\\
\texttt{mufti.mahmud@kfupm.edu.sa}
}

\begin{document}

\maketitle

\begin{abstract}
Split learning enables collaborative model training by partitioning neural networks across clients and servers. However, improper split placement can lead to severe privacy leakage through intermediate representations. In this work, we propose a topology-guided framework for privacy-aware split learning based on the persistent Betti complexity of smashed activations. Through comprehensive layer-wise analysis, we show that privacy risk in split learning is highly non-uniform across layers and exhibits sharp transition regions that are not captured by architectural depth alone. In particular, feature inversion fidelity increases from negligible reconstruction to as high as 0.98 SSIM at deeper, privacy-critical split points. We further demonstrate that Betti complexity consistently identifies representation regimes associated with elevated feature-space privacy leakage across architectures and datasets. Leveraging this observation, we introduce BettiSafe, a topology-guided split selection strategy that identifies privacy-sensitive layers without requiring explicit attack execution. BettiSafe improves resistance to feature inversion by 2–5$\times$ compared to depth-based heuristics while preserving classification accuracy. In addition, Betti-based regularisation increases inversion difficulty by nearly 5$\times$ without degrading model utility, enabling a favourable privacy–utility trade-off. Overall, our results highlight topological complexity as a promising structural descriptor for secure, adaptive, and representation-aware split learning in real-world collaborative systems
\end{abstract}

\section{Introduction}

Collaborative machine learning has emerged as a critical paradigm for enabling resource-constrained devices and data providers to jointly train deep models without centralising sensitive data.  While Federated Learning (FL) has become a widely adopted privacy-preserving training mechanism, split learning (SL) is increasingly gaining attention as an alternative. In the standard SL frameworks, a client computes a sequence of early neural layers, transmits the resulting intermediate representation (IR) to the server, and receives gradients during backpropagation. This architecture enables most computation to be offloaded to a powerful server while keeping raw data entirely on the client side, making SL attractive for privacy-sensitive domains such as healthcare, mobile sensing, and large-scale IoT systems.

Unlike FL, which has undergone considerable research in privacy aspects~\cite{9794622}, SL is still in its initial stages. Thus, apart from privacy via data locality and gradient smashing, SL provides a limited privacy guarantee by design. Recent studies have demonstrated that smashed activations exchanged at the split point may still leak substantial private information through reconstruction and inference attacks~\cite{ZENG2025107150}. Consequently, the privacy of SL fundamentally depends on \textbf{where} the model is partitioned.

However, selecting the split layer remains an open and underexplored problem. Most existing systems apply architectural heuristics (e.g., avoiding early layers) or communication-based constraints. While these approaches may reduce computation or bandwidth costs, they do not offer an architecture-agnostic mechanism for quantifying privacy leakage at a given split point. Moreover, evaluating candidate splits using reconstruction attacks is computationally expensive and impractical during deployment. As a result, current SL deployments lack a systematic method of identifying privacy-sensitive layers before leakage occurs. 

Motivated by these findings, this study introduces a topological perspective on privacy leakage in SL. The study initially observes that intermediate neural activations form geometric manifolds whose structure evolves across network depth. Early layers tend to produce low-dimensional representations closely tied to the input domain, while deeper layers encode increasingly abstract and compressed features. This observation motivates the hypothesis that the geometric and topological structure of IR is indicative of its attack susceptibility.


To quantify this structure, we leverage tools from topological data
analysis (TDA), specifically persistent homology and Betti numbers,
to characterise the geometry of smashed activations. We focus on the
first Betti number (Betti-1) as a descriptor of representation
complexity and derive a layer-wise privacy indicator, referred to as
the BettiLeak score, for ranking candidate split layers based solely
on activation structure.

Our key insight is that layers whose activations exhibit higher topological complexity tend to retain more reconstructible information about the original input, making them more vulnerable to privacy attacks. This relationship allows topological complexity to serve as a practical indicator of privacy risk in SL.

Building on this insight, we propose \textbf{BettiSplit}, a topology-guided framework for privacy-aware SL. BettiSplit comprises three
components: \textbf{BettiLeak}, a geometry-driven privacy indicator,
\textbf{BettiSafe}, a split-layer selection mechanism that ranks candidate split points based on estimated privacy risk and an optional Betti-based regularisation strategy that suppresses excessive topological complexity during training. Importantly, BettiSplit operates as a design-time framework and does not require executing privacy attacks during split selection.

We empirically evaluate our approach across multiple architectures, datasets, and threat models. The results demonstrate that topological complexity correlates strongly with privacy leakage and that Betti-guided split selection significantly reduces reconstruction risk compared to heuristic baselines while preserving model utility. Beyond its empirical contributions, this work establishes a conceptual bridge between topological analysis and privacy in collaborative learning. Rather than proposing new attacks or relying on heuristics, BettiSplit positions persistent homology as a diagnostic and decision-making tool for privacy-aware SL.

\subsection{Contributions}
This work makes the following contributions:
\begin{enumerate}
    \item \textbf{Topology-based characterisation of privacy leakage in SL:} We present a systematic study showing that the topological complexity of IR, quantified using persistent Betti numbers, correlates strongly with privacy leakage in SL.

    \item \textbf{BettiLeak: A geometry-driven privacy indicator for split layers}  We introduce BettiLeak, a lightweight metric derived from persistent homology that estimates the privacy sensitivity of IR directly from their topological structure. 

    \item \textbf{BettiSafe: topology-guided split-layer selection.} Building on BettiLeak, we propose BettiSafe, a split-layer selection mechanism {that identifies privacy-sensitive layers and improves
resistance against privacy attacks compared to depth-based heuristics.}
    
\item \textbf{Betti-based regularisation for suppressing topological leakage:} We explore the use of Betti-based penalties as a regularisation mechanism for reducing excessive topological complexity in smashed activations during training, improving resistance to reconstruction attacks without degrading task performance.


\item \textbf{Comprehensive empirical validation:}
We evaluate the proposed framework across multiple architectures, datasets, split locations, and attack settings, demonstrating the effectiveness of topology-guided privacy analysis and split selection under diverse learning scenarios.

\end{enumerate}

\section{Related Work}
\label{sec:related_work}
This section reviews prior work relevant to the proposed framework, focusing on four research directions: SL systems, privacy and security risks in SL, split-layer selection strategies, and the use of TDA in deep learning. Together, these studies highlight both the architectural foundations of SL and its emerging privacy challenges.

\subsection{Split Learning and Collaborative Training}

Existing split learning methods face scalability, privacy, and non-IID performance challenges. SplitBud provides one of the initial unified framework that modularises client, server, and strategy components for systematic evaluation of SL variants~\cite{10.1145/3721146.3721936}, but suffers from sequential training inefficiencies and reduced robustness under heterogeneous data. Cluster-based Parallel Split Learning addresses scalability through cluster-based parallel training and joint optimisation of cut-layer placement, clustering, and communication resources~\cite{10040976}, improving efficiency in resource-constrained environments while leaving privacy and personalisation largely unaddressed.


Several works combine SL and FL to address privacy and data heterogeneity. Privacy-Preserving Split Federated Learning (PPSFL) uses CNN-aware model partitioning and client-side Group Normalisation to improve personalisation and reduce information leakage~\cite{ZHENG2024231}. Another collaborative training framework, Dynamic Network Collaborative Learning investigates vertically partitioned training under device failures and network dynamics, improving reliability through fault-aware training and gossip-based ensembling~\cite{10.1145/3704413.3764450}. However, these approaches primarily target performance and scalability, offering limited mechanisms for quantifying or mitigating privacy leakage associated with split-layer placement.

\subsection{Privacy and Security Risks in Split Learning}
\label{subsec:sl_privacy_security}

Even though SL was originally proposed as a privacy-preserving collaborative learning paradigm, IRs, gradients, feature embeddings, and server-side model dynamics can leak substantial private information. Leveraging this, Sustainable Inference Attack incorporates the learning objective into the attack optimisation, enabling stealthy and persistent leakage throughout training, even under non-IID and vertically partitioned settings~\cite{YU2024396}. Similarly, utilising feature-space representation leakage, GAN-based reconstruction attacks recover high-fidelity data with minimal auxiliary knowledge, identifying feature alignment as a critical vulnerability~\cite{ZENG2025107150}.

Gradient-based attacks represent another major leakage channel. The EXACT~\cite{qiu2024evaluatingprivacyleakagesplit} attack reconstructs private features and labels by matching gradients at the cut layer, demonstrating that gradient sharing alone can expose sensitive information. Furthermore, the pseudo-client attack exploits intermediate server models for data reconstruction, model stealing, and label inference without access to client data or architectures~\cite{285505}. These findings indicate that privacy leakage is inherently embedded in the SL training process.

Recent studies further show that passive adversaries can infer sensitive information while remaining protocol-compliant, whereas malicious participants may compromise model integrity through poisoning attacks \cite{zhu2025passiveinferenceattackssplit,rieger2025safesplit}. Existing defences, including differential privacy~\cite{zhang2025model} and dedicated security mechanisms,  largely address individual attack classes, leaving privacy-aware split-layer selection unresolved.

\subsection{Split Point Selection Strategies}
Split-layer selection directly affects computation distribution, communication overhead, convergence behaviour, energy consumption, and privacy leakage in split and split-federated learning systems. Consequently, substantial effort has been put into understanding how split placement affects system performance and how it can be optimised under heterogeneous and resource-constrained environments.

Article~\cite{dachille2024impactcutlayerselection} highlights that the impact of cut layer selection varies across SFL, ranging from negligible effects on training behaviour to strong sensitivity under non-IID data distributions. To explicitly optimise split placement, \cite{10897845} formulates cut layer selection as a min–max latency optimisation problem in edge-assisted SFL. By modelling the relationship between split depth, computation cost, and communication overhead, the problem is relaxed into a tractable continuous form and solved via alternating optimisation. 

Adaptive approaches dynamically adjust split-layer placement during training. Dynamic Split Federated Learning (DSFL)~\cite{WAZZEH2025108275} employs a learned predictor and genetic algorithms to adapt cut-layer selection under heterogeneous resource conditions, mitigating straggler effects at the expense of additional overhead. Similarly, GA-Split~\cite{10.1145/3733566.3734433} uses evolutionary search to identify latency-efficient split configurations, although its focus is primarily on inference-time deployment. Other studies investigate the trade-off between energy consumption and privacy across different split depths~\cite{10516589}, showing that deeper splits increase client energy costs while reducing susceptibility to smashed-data reconstruction attacks. In summary, existing methods primarily optimise system-level metrics, with limited focus on structural indicators of privacy risk.

\subsection{Topological Data Analysis for Deep Learning}
\label{subsec:tda_literature}

Topological Data Analysis (TDA) provides a framework for characterising global and noise-robust structure in high-dimensional data through tools such as persistent homology. Recent surveys on Topological Deep Learning (TDL) highlight the integration of topological descriptors with neural representations to improve interpretability and robustness~\cite{zia2023topological,su2025topological}. These works emphasise a fundamental tradeoff: deeper models offer strong learning capacity but lack guarantees on global structure, whereas TDA is theoretically grounded but not inherently learnable. Building on this foundation, topology has been proposed as a core representational domain for relational learning rather than merely an auxiliary analysis tool~\cite{papamarkou2024position}.

Several studies employ TDA for for post hoc analysis of trained models. In~\cite{BALLESTER2024127787}, persistent homology is used to characterise functional connectivity structures and predict generalisation behaviour without test data. GraphPulse extends TDA to temporal graphs through mapper-based topological summaries and sequential neural models~\cite{shamsi2024graphpulse},achieving improved prediction performance on temporal networks~\cite{shamsi2024graphpulse}. While these works demonstrate the value of topology for representation analysis, they focus primarily on interpretability and graph learning. The use of topological descriptors to characterise privacy leakage in collaborative learning remains largely unexplored.


\subsection{Positioning of This Work}
\label{subsec:positioning}



Existing research has extensively studied privacy risks and split-point trade-offs in SL/SFL, while advances in TDA have demonstrated the value of topology-aware representations for capturing global structural properties. However, the use of topological information to characterise privacy leakage and guide architectural decisions in collaborative learning remains largely unexplored.

This work bridges these domains by investigating topology-informed descriptors of IRs as indicators of information leakage and as a basis for split-layer selection. Rather than relying solely on heuristic or system-level optimisation criteria, we leverage the global structure of neural representations to support privacy-aware architectural decision-making in SL.

\begin{algorithm}[htbp]
\caption{\textsc{BettiLeak} + \textsc{BettiSafe}: Topology-Guided Split Selection (Attack-Free)}
\label{alg:bettileak_bettisafe}
\begin{algorithmic}[1]
\REQUIRE Network $f = f_L \circ \cdots \circ f_1$ with $L$ layers; calibration mini-batches $\{x^{(b)}\}_{b=1}^{B}$; TDA limits $N_{\max}$ (max samples), $D_{\max}$ (max feature dim); trade-off $\lambda \ge 0$.
\ENSURE Recommended split index $k^\star \in \{1,\dots,L-1\}$; BettiLeak scores $\{\beta_k\}_{k=1}^{L-1}$.

\STATE Initialise empty list $\beta \leftarrow [\;]$

\STATE \textbf{Procedure} \textsc{BettiLeakScore:}$(Z, N_{\max}, D_{\max})$
\STATE \hspace{0.6cm} Flatten $Z$ to matrix $A \in \mathbb{R}^{N \times D}$
\STATE \hspace{0.6cm} \textbf{if} $N > N_{\max}$ \textbf{then} subsample rows of $A$ to $N_{\max}$ \textbf{end if}
\STATE \hspace{0.6cm} \textbf{if} $D > D_{\max}$ \textbf{then} subsample columns of $A$ to $D_{\max}$ \textbf{end if}
\STATE \hspace{0.6cm} Compute persistence diagrams $\mathrm{dgms} \leftarrow \mathrm{Ripser}(A)$
\STATE \hspace{0.6cm} Let $\mathrm{dgms}_1$ be the 1-dimensional diagram (loops)
\STATE \hspace{0.6cm} $\beta_1 \leftarrow \left|\{(b,d)\in \mathrm{dgms}_1 : d>b\}\right|$
\STATE \hspace{0.6cm} \textbf{return} $\beta_1 / (N \cdot D)$

\FOR{$k = 1$ \textbf{to} $L-1$}
    \STATE Initialise empty container $S \leftarrow [\;]$
    \FOR{$b = 1$ \textbf{to} $B$}
        \STATE $z^{(b)} \leftarrow (f_k \circ \cdots \circ f_1)(x^{(b)})$ \COMMENT{smashed activation at cut $k$}
        \STATE Append $z^{(b)}$ to $S$
    \ENDFOR
    \STATE $Z_k \leftarrow \mathrm{Concat}(S)$
    \STATE $\beta_k \leftarrow \textsc{BettiLeakScore}(Z_k, N_{\max}, D_{\max})$
    \STATE Append $\beta_k$ to list $\beta$
\ENDFOR

\STATE Normalise $\{\beta_k\}_{k=1}^{L-1}$ to $\{\hat{\beta}_k\}_{k=1}^{L-1} \subset [0,1]$

\IF{$\lambda = 0$}
    \STATE $k^\star \leftarrow \arg\min_{k \in \{1,\dots,L-1\}} \hat{\beta}_k$
\ELSE
    \FOR{$k = 1$ \textbf{to} $L-1$}
        \STATE $c_k \leftarrow \sum_{i=1}^{k} \hat{\beta}_i$ \COMMENT{client cumulative complexity}
        \STATE $s_k \leftarrow \sum_{i=k+1}^{L} \hat{\beta}_i$ \COMMENT{server cumulative complexity}
        \STATE $J_k \leftarrow s_k - \lambda \, c_k$
    \ENDFOR
    \STATE $k^\star \leftarrow \arg\max_{k \in \{1,\dots,L-1\}} J_k$
\ENDIF

\RETURN $k^\star$, $\{\beta_k\}_{k=1}^{L-1}$
\end{algorithmic}
\end{algorithm}

\begin{algorithm}[htbp]
\caption{\textsc{Betti-Reg}: Betti-Regularised Training to Suppress Topological Leakage}
\label{alg:betti_reg_training}
\begin{algorithmic}
\REQUIRE Network $f = f_L \circ \cdots \circ f_1$; training data loader $\mathcal{D}$; chosen split (penalty) layer index $k$; learning rate $\eta$; Betti weight $\lambda_B \ge 0$; TDA limits $N_{\max}, D_{\max}$.
\ENSURE Trained parameters $\theta$.

\STATE Initialise parameters $\theta$
\FOR{each mini-batch $(X,Y) \sim \mathcal{D}$}
    \STATE Forward pass: $H_0 \leftarrow X$
    \FOR{$i = 1$ \textbf{to} $L$}
        \STATE $H_i \leftarrow f_i(H_{i-1})$
        \IF{$i = k$}
            \STATE $Z \leftarrow H_i$ \COMMENT{smashed activations used for Betti penalty}
        \ENDIF
    \ENDFOR
    \STATE $\hat{Y} \leftarrow H_L$
    \STATE $\mathcal{L}_{\mathrm{CE}} \leftarrow \mathrm{CE}(\hat{Y}, Y)$

    \STATE \COMMENT{Betti penalty on smashed activations}
    \STATE Flatten $Z$ to $A \in \mathbb{R}^{N \times D}$
    \STATE \textbf{if} $N > N_{\max}$ \textbf{then} subsample rows of $A$ \textbf{end if}
    \STATE \textbf{if} $D > D_{\max}$ \textbf{then} subsample columns of $A$ \textbf{end if}
    \STATE $\mathrm{dgms} \leftarrow \mathrm{Ripser}(A)$; let $\mathrm{dgms}_1$ be 1-D diagram
    \STATE $\mathcal{L}_{\mathrm{B}} \leftarrow \frac{\left|\{(b,d)\in \mathrm{dgms}_1 : d>b\}\right|}{N \cdot D}$

    \STATE Total loss: $\mathcal{L} \leftarrow \mathcal{L}_{\mathrm{CE}} + \lambda_B \, \mathcal{L}_{\mathrm{B}}$
    \STATE Update: $\theta \leftarrow \theta - \eta \nabla_{\theta}\mathcal{L}$
\ENDFOR

\RETURN $\theta$
\end{algorithmic}
\end{algorithm}
 
\section{Methodology}
\label{sec:methodology}

This section presents \textit{BettiSplit}, a topology-guided framework for privacy-aware SL. The system comprises three components: (i) \textit{BettiLeak}, a topology-based privacy leakage metric derived from persistent homology, (ii) \textit{BettiSafe}, a privacy-aware split-layer selection mechanism, and (iii) a Betti-regularised training strategy. Together, these components support both design-time split selection and deployment-time privacy-aware training.

\subsection{System Overview}

Consider a neural network partitioned at layer $k$ into a client-side subnetwork $f_{0:k}$ and a server-side subnetwork $f_{k+1:L}$. Given an input $x$, the client computes the IRs 

\[
h_k = f_{0:k}(x),
\]

which is transmitted to the server. The server completes the forward pass and produces the prediction

\[
\hat{y} = f_{k+1:L}(h_k).
\]

The choice of split layer $k$ determines the exchanged IR and therefore influences both privacy exposure and system behaviour. BettiSplit analyses the topology of IRs across candidate split layers to estimate privacy risk and guide split-layer selection. At a high level, the framework first quantifies the topological complexity of smashed activations and subsequently selects a split layer using the resulting privacy indicator. The complete procedure is formalised in Algorithm~\ref{alg:bettileak_bettisafe}.
\subsection{Topological Characterisation of Neural Representations}
\label{subsec:topological_characterisation}
Since smashed activations constitute the primary information exchanged in SL, we analyse their topology to obtain a representation-level
characterisation of privacy-relevant structure. Given a calibration mini-batch, the smashed activations at split layer $k$ are represented by

\[
h_k \in \mathbb{R}^{N \times d},
\]

where $N$ denotes the number of samples and $d$ the feature dimension. Each activation vector is treated as a point in feature space, forming a point cloud that characterises the geometry of the intermediate representation.

To analyse the global structure of this point cloud, we employ persistent homology using a Vietoris--Rips filtration~\cite{koyama2024distilledvietorisripsfiltration}. Persistent homology tracks the appearance and disappearance of topological features across multiple spatial scales and summarises them through persistence diagrams. Unlike constructions that require explicit manifold assumptions, Vietoris--Rips complexes operate directly on pairwise distances, making them suitable for high-dimensional neural representations whose underlying geometry is generally unknown.

The resulting topological features are quantified using Betti numbers. In particular, the $0$-dimensional Betti number $\beta_0$ counts connected components, whereas the $1$-dimensional Betti number $\beta_1$ counts loop structures within the activation space. While $\beta_0$ often stabilises rapidly as representations become connected, $\beta_1$ remains sensitive to non-linear feature interactions induced by hierarchical composition in deep networks.

Accordingly, this work uses $\beta_1$ as the primary descriptor of representation complexity. For a persistence diagram $\mathrm{dgms}_1$ corresponding to the one-dimensional homology group, the associated loop-level complexity is given by

\[
\beta_1
=
\left|
\left\{
(b,d)\in\mathrm{dgms}_1 : d>b
\right\}
\right|,
\]

where $(b,d)$ denotes the birth and death scales of a persistent one-dimensional topological feature. This quantity forms the basis of the topology-guided privacy metric introduced in the following subsection.
\subsection{Betti Leakage Metric}
\label{subsec:betti_leakage}

Feature inversion attacks seek to recover an input from its
corresponding smashed activation. Since reconstruction behaviour
depends on the structure of the underlying representation space,
we investigate whether the loop-level topological complexity
captured by $\beta_1(h_k)$ can serve as an indicator of privacy
exposure. Higher values of $\beta_1(h_k)$ correspond to richer
representation structure and potentially greater susceptibility
to inversion-based attacks. Accordingly, we define the Betti leakage score as:

\[
\textbf{\text{BettiLeak}(k)} =
\frac{\left|\{(b,d)\in \mathrm{dgms}_1 : d>b\}\right|}{N \cdot D}.
\]

BettiLeak is gradient-free, architecture-agnostic, and computable using only forward activations, forming the core privacy indicator used in Algorithm~\ref{alg:bettileak_bettisafe} for split-layer selection.

\subsection{BettiSafe Split Selection}
\label{subsec:bettisafe}

BettiSafe is a privacy-aware split selection mechanism that ranks candidate split layers using the BettiLeak score. For each candidate layer $k$, BettiLeak is evaluated on calibration mini-batches and normalised across all feasible split locations to obtain $\hat{\beta}_k$.

When privacy is the sole objective, BettiSafe selects the split layer exhibiting the lowest normalised topological leakage:

\[k^\star=\arg\min_{k\in\mathcal K}
\hat{\beta}_k\]

where $\mathcal K$ denotes the set of feasible split layers. To account for deployment constraints, BettiSafe can additionally balance client-side and server-side topological complexity. Let:

\[
c_k=\sum_{i=1}^{k}\hat{\beta}*i~~\text{and}~~s_k=\sum*{i=k+1}^{L}\hat{\beta}_i
\]

denote the cumulative client and server complexities, respectively. The split layer is then selected according to

\[
J_k=s_k-\lambda c_k~~\text{and}~~k^\star=\arg\max_{k\in\mathcal K}J_k,
\]

where $\lambda$ controls the trade-off between privacy-oriented topology reduction and computational distribution across the split architecture. Unlike attack-based evaluation procedures, BettiSafe operates using only forward activations and does not require model retraining, reconstruction attacks, or access to raw client data.
 
\subsection{Betti-Regularised Split Learning}
\label{subsec:betti_regularisation}

While BettiSafe identifies privacy-efficient split locations, the selected IRs may still  retain non-trivial topological complexity. To further suppress privacy-relevant structure, we introduce a Betti-based regularisation term during training.

Let $\mathcal{L}_{\mathrm{CE}}$ denote the standard classification loss. We define the total training objective as

\[
\mathcal{L}_{\mathrm{total}} =
\mathcal{L}_{\mathrm{CE}} + \lambda \cdot \beta_1(h_k),
\]

where $\beta_1(h_k)$ denotes the normalised loop-level topological complexity of the smashed activations at the selected split layer, and $\lambda$ controls the strength of regularisation.

By penalising persistent loop structures within the activation space, the proposed regulariser encourages simpler and less input-specific intermediate representations while preserving task performance. The corresponding training procedure is formalised in Algorithm~\ref{alg:betti_reg_training}.

\begin{figure}
    \centering
    \includegraphics[height=7cm,width=.7\linewidth]{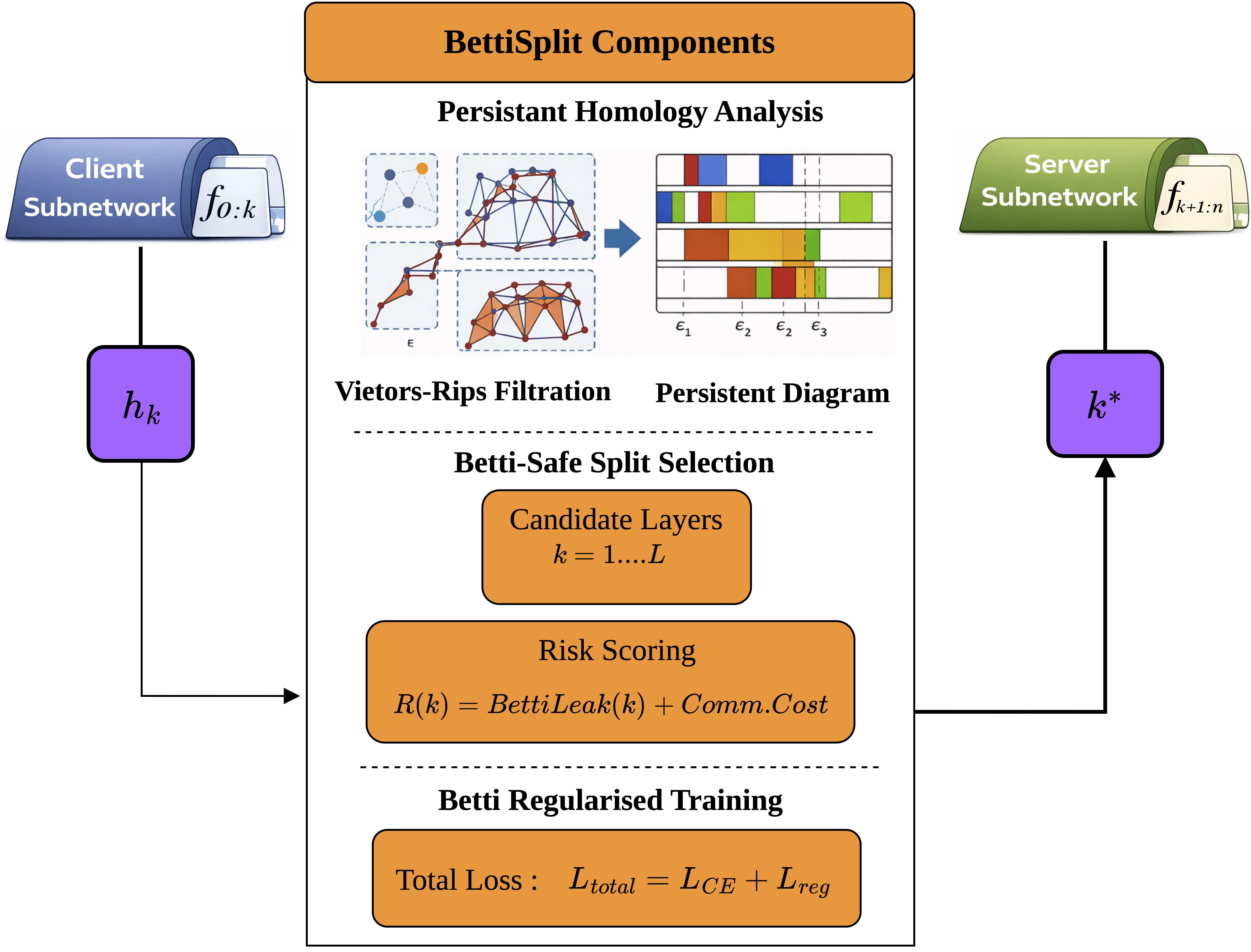}
    \caption{The workflow diagram of the BettiSplit framework}
    \label{fig1}
\end{figure}

The overall workflow of the proposed BettiSplit framework is illustrated in Fig.~\ref{fig1}.

\section{Security Analysis}
This section analyses the privacy implications of SL under the proposed BettiSplit framework. We first define the threat model and then review reconstruction attacks applicable to SL. Finally, we discuss how the topological complexity of smashed activations influences attack feasibility. Empirical evaluation is presented in
Section~\ref{sec:results}.
\subsection{Threat Model }
We consider the standard honest-but-curious server setting commonly adopted in SL security literature. The client holds raw data $x$ and computes the first $k$ layers of the network, while the server receives smashed activations $h_k = f_{0:k}(x)$ and performs the remaining forward and backward computations.

The server follows the training protocol but may attempt to infer sensitive information from exchanged activations or gradients. The
adversary is assumed to have access to:
\begin{itemize}
    \item Smashed activations $h_k$,
    \item Gradients with respect to server-side parameters during training,
    \item Full knowledge of the model architecture and parameters,
    \item The ability to execute optimisation-based reconstruction attacks.
\end{itemize}

The study does not consider active adversaries, protocol deviations, or collusion between clients. This threat model reflects practical deployments where computation is outsourced to powerful but untrusted servers.

\subsection{Gradient-Based Reconstruction Attacks}

Gradient-based attacks attempt to reconstruct private inputs by exploiting gradients shared during training. Deep Leakage from Gradients~\cite{NEURIPS2019_60a6c400} reconstructs inputs by optimising a dummy sample whose gradients match the true gradients observed by the server:

\[
\tilde{x}^\star = 
\arg\min_{\tilde{x}}
\left\|
\nabla_\theta \mathcal{L}(f(\tilde{x})) - \nabla_\theta \mathcal{L}(f(x)) \right\|_2^2.
\]

In SL, the server observes gradients only for the server-side subnetwork, limiting the information available to gradient-based attacks compared to centralised or FL settings. Leakage is evaluated using reconstruction error, measured by mean squared error (MSE) between reconstructed and original inputs.

An improved variant, iDLG~\cite{zhao2020idlgimproveddeepleakage}, leverages structural properties of gradients to infer ground-truth labels prior to input reconstruction. While effective when gradients of the final classification layer are exposed, its applicability in SL is limited as the server typically observes gradients only for server-side layers. However, we include iDLG to account for label-assisted gradient attacks under early split configurations.

\subsection{Feature Inversion Attacks}
Feature inversion attacks~\cite{Mahendran_2015_CVPR} operate at inference time and attempt to reconstruct inputs directly from smashed activations by solving:
\[
\tilde{x}^\star = \arg\min_{\tilde{x}} \left\| f_{0:k}(\tilde{x}) - h_k
\right\|_2^2.
\]

The success of feature inversion depends on the amount of input-specific structure preserved in IR. When smashed activations retain rich geometric or spatial structure, inversion attacks can recover high-fidelity approximations of the original input. Conversely, more compressed and entangled representations restrict the feasible reconstruction space and reduce attack success.

We evaluate the effectiveness of feature inversion using reconstruction error metrics, including MSE and structural similarity index (SSIM), which capture both fidelity and perceptual similarity between the reconstructed and original inputs.


\subsection{Topological Complexity and Attack Feasibility}

The proposed BettiLeak metric characterises loop-level topological structure in smashed activations. Since reconstruction attacks operate directly on the exchanged representation space, the organisation of these representations influences attack feasibility.

Representations exhibiting greater topological complexity may retain richer input-specific structure, potentially facilitating reconstruction-based attacks. Conversely, more topologically compact representations are expected to reduce reconstruction fidelity. This motivates the use of persistent homology as an attack-agnostic indicator of privacy sensitivity in SL. The validity of this hypothesis is evaluated empirically through gradient-based and feature inversion attacks in Section~\ref{sec:results}.

\subsection{BettiSafe and Betti-Regularisation as Defences}
BettiSafe mitigates privacy risk by selecting split layers whose smashed activations exhibit reduced topological complexity, without requiring execution of reconstruction attacks during split selection. This design-time selection limits exposure to privacy-sensitive representations.

In addition, Betti-based regularisation suppresses excessive topological structure in smashed activations during training. By discouraging the formation of loop-level features associated with input-specific structure, this mechanism provides an auxiliary defence that complements topology-guided split selection.

Overall, BettiSplit frames split-layer selection and training as security-aware decisions grounded in the topology of IRs, rather than heuristic or attack-specific considerations.

\section{Results}
\label{sec:results}
This section presents an empirical evaluation of the proposed topology-guided framework for privacy-aware SL. The objective of the analysis is to characterise how privacy leakage evolves across split locations and to evaluate whether topological descriptors of IRs provide a reliable and generalisable signal for identifying privacy-critical layers.

Unless stated otherwise, all experiments follow a standard SL protocol with a fixed client-server partition, where only the split layer varies. Models are trained under identical optimisation settings and evaluated using consistent attack configurations. Throughout this section, the primary objective is to characterise privacy leakage and identify privacy-critical representation regions rather than optimise computational performance.

\subsection{Experimental Scope}

We begin with a LeNet-style convolutional neural network trained on the FashionMNIST, which serves as the canonical experimental setting for layer-wise privacy analysis. Privacy leakage is evaluated across all candidate split layers using feature inversion and gradient-based reconstruction attacks (DLG). The topological complexity of IRs is quantified using normalised Betti-1 values computed from smashed activations.

To assess robustness and generalisation, the analysis is extended to multiple architectures, including SimpleCNN, ResNet-18, and MobileNetV2, and
additional datasets such as CIFAR-10 and SVHN. These experiments evaluate whether topology-guided privacy indicators remain informative under architectural and data-distribution shifts.

\subsection{Layer-wise Privacy Leakage Across Split Points}
\label{subsec:layerwise_privacy}

We first examine how privacy leakage evolves as a function of the split layer. For a fixed model architecture, the cut location is progressively moved from early to deeper layers, and the privacy risk associated with each split is evaluated.

Feature inversion attacks reconstruct the original input from smashed activations shared at the split layer. Reconstruction quality is measured using the SSIM, where higher values indicate greater privacy leakage. In addition, gradient-based attacks (DLG) are evaluated to assess leakage through shared gradients.

Figure~\ref{fig:ssim-layerwise} reports SSIM-based inversion leakage across all candidate split layers for LeNet on FashionMNIST. Early split points (layers 0--5) exhibit negligible reconstruction quality (SSIM $\approx 0$), indicating strong resistance to feature inversion. Beyond layer~6, reconstruction quality increases abruptly, revealing a distinct privacy transition in inversion vulnerability.


This behaviour indicates the presence of a privacy transition region within the neural network, where specific IRs retain sufficient information to enable reconstruction while still being exposed to the server. Notably, neighbouring layers can exhibit substantially different privacy behaviour, indicating that depth alone is insufficient for identifying privacy-critical split locations.

Table~\ref{tab:ssim-layer} summarises the corresponding SSIM leakage, DLG reconstruction error, and representation statistics across all candidate split layers. Variance and spectral entropy are included as conventional non-topological descriptors of activation distributions for comparison with the proposed topology-based metric.

While early layers exhibit uniformly low inversion leakage, reconstruction quality increases substantially within the transition region. In contrast, DLG reconstruction error remains relatively stable across split locations, even where feature inversion becomes feasible. This divergence suggests that feature-space and gradient-based attacks exploit different properties of IRs and motivates the need for attack-specific privacy characterisation.


\begin{table}[!b]
\centering
\caption{Layer-wise SSIM leakage and corresponding representation metrics. 
Betti-1 measures topological complexity, while variance and spectral entropy 
represent conventional statistical descriptors of the activation distribution.}
\label{tab:ssim-layer}
\begin{tabular}{c c c c c c}
\hline
\textbf{Layer} & \textbf{Betti-1} & \textbf{Variance} & \textbf{Entropy} & \textbf{SSIM} & \textbf{DLG Error} \\ \hline
0  & 0.001465 & 0.24 & 4.13 & 0.000000 & 1.215659 \\
1  & 0.002197 & 0.24 & 4.13 & 0.000000 & 1.322053 \\
2  & 0.001953 & 0.24 & 4.13 & 0.000000 & 1.064225 \\
3  & 0.002075 & 2.47 & 4.29 & 0.000000 & 1.253145 \\
4  & 0.002563 & 2.47 & 4.29 & 0.000000 & 1.332093 \\
5  & 0.001953 & 1.50 & 4.08 & 0.000000 & 1.070536 \\
6  & 0.002197 & 1.50 & 4.08 & 0.801544 & 1.135165 \\
7  & 0.001563 & 1.50 & 4.08 & 0.594194 & 1.249030 \\
8  & 0.002344 & 5.09 & 3.59 & 0.911887 & 1.057114 \\
9  & 0.003534 & 5.09 & 3.59 & 0.912519 & 1.156925 \\
10 & 0.002790 & 5.09 & 3.59 & 0.982561 & 1.070255 \\
11 & 0.023438 & 5.09 & 3.59 & 0.985871 & 1.156663 \\ \hline
\end{tabular}
\end{table}

\subsection{Topological Complexity and Privacy Leakage}
\label{subsec:topology_privacy}

The study next examines whether the observed layer-wise variation in privacy leakage can be explained through topological properties of intermediate representations. Specifically, we analyse the relationship between Betti-1 complexity of smashed activations and feature inversion leakage.

For each candidate split layer, persistent homology is computed over the activation point cloud obtained from a fixed minibatch of inputs. The resulting Betti-1 values quantify the number of loop-level topological features persisting across scales, capturing non-linear geometric structure in the representation. For clarity, the reported Betti-1 values form the basis of the BettiLeak metric used for split-layer selection.

Figure~\ref{fig:betti-layerwise} shows the evolution of Betti-1 values across network depth. Betti-1 remains relatively low and stable throughout the early convolutional layers, followed by a noticeable increase in deeper layers. This increase aligns closely with the emergence of inversion vulnerability observed in the previous subsection.

\begin{figure*}[!t]
\centering

\begin{minipage}[t]{0.32\textwidth}
\centering
\begin{tikzpicture}
\begin{axis}[
xlabel={Split Layer},
ylabel={SSIM},
width=\linewidth,
height=0.65\linewidth,
ymin=0,
ymax=1,
grid=major,
]
\addplot+[mark=square*] coordinates {
(0,0.0)
(1,0.0)
(2,0.0)
(3,0.0)
(4,0.0)
(5,0.0)
(6,0.801544)
(7,0.594194)
(8,0.911887)
(9,0.912519)
(10,0.982561)
(11,0.985871)
};
\end{axis}
\end{tikzpicture}
\caption{Layer-wise SSIM reconstruction leakage across split layers.}
\label{fig:ssim-layerwise}
\end{minipage}
\hfill
\begin{minipage}[t]{0.32\textwidth}
\centering
\begin{tikzpicture}
\begin{axis}[
xlabel={Layer Index},
ylabel={Betti-1},
width=\linewidth,
height=0.65\linewidth,
ymin=0,
grid=major,
]
\addplot+[mark=*] coordinates {
(0,0.00134277)
(1,0.00128174)
(2,0.00146484)
(3,0.00152588)
(4,0.00164795)
(5,0.00134277)
(6,0.00177002)
(7,0.00286458)
(8,0.00169271)
(9,0.00260417)
(10,0.00297619)
(11,0.015625)
};
\end{axis}
\end{tikzpicture}
\caption{Layer-wise Betti-1 complexity across the LeNet architecture.}
\label{fig:betti-layerwise}
\end{minipage}
\hfill
\begin{minipage}[t]{0.32\textwidth}
\centering
\begin{tikzpicture}
\begin{axis}[
xlabel={Split Layer},
ylabel={SSIM},
width=\linewidth,
height=0.65\linewidth,
ymin=-0.05,
ymax=1.05,
grid=major,
]
\addplot+[mark=square*] coordinates {
(0,1.0000)
(1,1.0000)
(2,0.9996)
(3,0.7007)
(4,0.2279)
(5,0.0724)
(6,0.0695)
(7,0.0461)
(8,0.0358)
(9,0.0210)
(10,0.0245)
(11,-0.0049)
(12,0.0056)
(13,0.0012)
};
\end{axis}
\end{tikzpicture}
\caption{SSIM-based inversion leakage across split layers on CIFAR-10 (SimpleCNN).}
\label{fig:cifar-ssim-layerwise}
\end{minipage}

\label{fig:layerwise_privacy_topology}

\end{figure*}
Figure~\ref{fig:betti-ssim-crossmodel} and~\ref{fig:phase_diagram_topology_leakage} plots Betti-1 values against SSIM-based inversion leakage across candidate split layers.

While the relationship is not strictly monotonic, layers exhibiting higher Betti-1 complexity tend to correspond to higher reconstruction quality. This association provides a structural explanation for the empirical privacy transition and is more informative than depth-based heuristics alone, which fail to identify privacy-critical regions. These observations support the hypothesis that representation geometry captured by topological complexity is associated with inversion-based privacy leakage in SL and can serve as a useful indicator of privacy-sensitive representation regions.

\begin{table}[t]
\centering
\caption{Correlation between Betti-1 complexity and inversion leakage (SSIM) across candidate split layers.}
\label{tab:betti_correlation}
\begin{tabular}{c c c}
\hline
Dataset / Model & Pearson $r$ & Spearman $\rho$ \\
\hline
FashionMNIST / LeNet & 0.842 & 1.000 \\
FashionMNIST / ResNet18 & 0.329 & -0.200 \\
FashionMNIST / MobileNetV2 & -0.714 & -0.800 \\
CIFAR-10 / MobileNetV2 & 0.781 & 0.800 \\
\hline
\end{tabular}
\end{table}

To further quantify this relationship, Table~\ref{tab:betti_correlation} reports Pearson and Spearman correlation coefficients between Betti-1 complexity and SSIM-based inversion leakage across candidate split layers for multiple architectures and datasets. Strong positive correlations are observed in several configurations,including LeNet on FashionMNIST ($r=0.842$) and MobileNetV2 on CIFAR-10 ($r=0.781$), indicating that increases in topological complexity generally align with higher reconstruction fidelity. 
However, the relationship is not strictly monotonic across all architectures. For example, MobileNetV2
on FashionMNIST exhibits a negative Pearson correlation despite displaying a clear alignment between increases in Betti complexity and
the onset of inversion vulnerability observed in
Fig.~\ref{fig:betti-ssim-crossmodel}. This behaviour suggests that Betti-based topology is most effective for identifying
privacy transition regions rather than acting as a universal linear predictor of reconstruction quality.

\subsection{Topology, Bottlenecks, and Inversion Vulnerability}
\label{subsec:bottleneck_case}

To further illustrate the relationship between representation topology and inversion susceptibility, we consider a controlled comparison between two LeNet-based models trained on FashionMNIST under identical optimisation settings: a bottlenecked variant with a reduced latent dimension and a wide variant with a standard latent layer. The bottleneck is introduced solely to alter the internal representation geometry while preserving task performance. The results is summarised in Table.~\ref{tab:bottleneck}.


\begin{table}[!b]
\centering
\caption{Controlled bottleneck experiment isolating topological complexity from classification accuracy. Both models share identical architectures except for the latent dimensionality at the final layer.}
\label{tab:bottleneck}
\begin{tabular}{l c c c c}
\hline
\textbf{Model Variant} & \textbf{Latent Dim} & \textbf{Accuracy (\%)} & \textbf{Betti-1} & \textbf{SSIM} \\
\hline
Wide LeNet        & 120 & 97.73 & 0.0230 & 0.656 \\
Bottlenecked LeNet & 20  & 96.54 & 0.0160 & 0.535 \\
\hline
\end{tabular}
\end{table}

Both models achieve comparable classification accuracy (96.54\% versus 97.73\%), indicating that neither model is severely underfitting and that predictive performance is not the primary driver of privacy leakage. Despite similar accuracy, clear differences emerge in the structure and privacy sensitivity of their latent representations.

The wide model exhibits higher topological complexity in its latent space, with a Betti-1 value of \textbf{0.0230} compared to \textbf{0.0160} for the bottlenecked model. This increase in loop-level topological features corresponds to a richer and less compressed representation geometry. Consistent with this observation, feature inversion attacks achieve higher reconstruction fidelity for the wide model (SSIM = \textbf{0.656}) than for the bottlenecked model (SSIM = \textbf{0.535}).

This controlled experiment isolates representation topology from predictive performance and demonstrates that increased topological complexity is associated with greater inversion vulnerability. The result reinforces the use of Betti-based descriptors as privacy-relevant indicators beyond conventional metrics such as classification accuracy or network depth. 


\subsection{Consistency Across Privacy Attack Models}
\label{subsec:attack_consistency}
Having established the relationship between topology and feature inversion leakage, we next examine whether similar behaviour is observed under gradient-based reconstruction attacks. For this, Deep Leakage from Gradients (DLG) is evaluated across all candidate split layers using the same experimental configuration.

Table~\ref{tab:ssim-layer} reports DLG reconstruction errors for each split location. In contrast to feature inversion, DLG reconstruction error remains relatively stable across network depth and does not exhibit the distinct privacy transition region observed in Fig.~\ref{fig:ssim-layerwise}. Furthermore, no meaningful correlation is observed between Betti-1 complexity and DLG reconstruction performance.

These results suggest that topology-based indicators are particularly informative for feature-space leakage in SL, where attacks operate directly on exchanged representations. In contrast, gradient-based leakage appears to be governed by optimisation dynamics and gradient structure rather than the geometric organisation of smashed activations. This distinction highlights the difference between inference-time and training-time attack surfaces and motivates the use of complementary privacy evaluation methodologies.

\subsection{Effect of Betti-Regularised Training}
\label{subsec:betti_reg}
The study further evaluates the impact of Betti-based regularisation on privacy and model utility. Table~\ref{tab:betti-reg} compares classification accuracy, topological complexity, and inversion error between baseline SL and Betti-regularised training.

Betti regularisation reduces the Betti-1 complexity of smashed activations from 0.023 to 0.015, indicating a substantial decrease in loop-level topological structure. This reduction is accompanied by a nearly five-fold increase in inversion error, suggesting that topologically simpler representations are more resistant to reconstruction attacks.

Importantly, the reduction in privacy leakage does not come at the expense of predictive performance. Classification accuracy remains comparable to the baseline and exhibits a slight improvement in this experiment. These results indicate that topology-aware regularisation can provide a favourable privacy--utility trade-off and serves as a useful complement to topology-guided split selection.

\begin{table}[htbp]
\centering
\caption{Effect of Betti regularisation on privacy and accuracy. Betti-1 values indicate the topological complexity of smashed activations.}
\label{tab:betti-reg}
\begin{tabular}{l c c c}
\hline
\textbf{Model} & \textbf{Accuracy (\%)} & \textbf{Betti-1} & \textbf{Inversion MSE} \\ \hline
Baseline          & 85.93 & 0.023 & $3.99 \times 10^{-4}$ \\
Betti-Regularised & 86.74 & 0.015 & $1.96 \times 10^{-3}$ \\ \hline
\end{tabular}
\end{table}
\subsection{Cross-Architecture Generalisation on FashionMNIST}
\label{subsec:cross_model}

To evaluate generalisation across architectures, we repeat the analysis on FashionMNIST using three distinct models: LeNet, SimpleCNN, and ResNet-18. 

Figure~\ref{fig:betti-ssim-crossmodel} plots Betti-1 complexity against SSIM-based inversion leakage across all split layers and architectures. Despite substantial architectural differences, all models exhibit a consistent structural trend: layers with higher Betti-1 complexity correspond to increased inversion leakage. For example, in the MobileNetV2–FashionMNIST configuration the BettiLeak scores across candidate layers 
$\{4,9,14,17\}$ were $\{0.0068, 0.0083, 0.0063, 0.0031\}$,
leading BettiSafe to select layer 17 as the privacy-aware split point.

Table~\ref{tab:fm-split-compare} summarises Betti-guided and heuristic split selection results across both FashionMNIST and CIFAR-10. The results indicate that the relationship between topological complexity and privacy leakage is not specific to a particular network design, but persists across architectural variations under a fixed data distribution. These findings support the architecture-agnostic nature of the proposed topology-guided privacy indicator.

\subsection{Cross-Dataset Generalisation}
\label{subsec:cross_dataset}

The study further evaluates whether topology-guided privacy characterisation generalises beyond FashionMNIST to more complex data distributions. Thus, the analysis is repeated on CIFAR-10 using LeNet-style CNN, SimpleCNN, and ResNet-18 architectures. Across all models, feature inversion leakage exhibits a distinct privacy transition region, although its location varies according to network architecture.

Unlike FashionMNIST, early split layers on CIFAR-10 permit higher-fidelity reconstruction due to the increased complexity and variability of the input distribution, resulting in elevated SSIM values at shallow split locations. Nevertheless, Figure~\ref{fig:cifar-ssim-layerwise} shows that increases in Betti-1 complexity continue to align with the onset of inversion vulnerability across architectures, indicating that topology remains informative despite the greater representational complexity of the dataset.

To further assess robustness under distribution shifts, the analysis is extended to SVHN. As summarised in Table~\ref{tab:fm-split-compare}, the relationship between topological complexity and inversion leakage becomes less evident than on FashionMNIST and CIFAR-10, reflecting the increased visual variability of the dataset. However, Betti-guided split selection remains competitive across privacy evaluation settings, suggesting that topological descriptors continue to provide useful privacy information even when the correlation between topology and reconstruction quality weakens.

Overall, these results indicate that the proposed topology-guided privacy indicators generalise across substantially different data distributions and are not tied to a specific dataset or image domain.

\subsection{Topology-Guided Split Selection}
\label{subsec:split_selection}

Having established the relationship between topological complexity and privacy leakage, we next evaluate whether Betti-based profiling can be used to guide split-layer selection in practice. Table~\ref{tab:fm-split-compare} compares Betti-guided split selection against a depth-based heuristic across multiple architectures and datasets. For each configuration, the selected split layer is evaluated using feature inversion (SSIM) and gradient-based reconstruction attacks, including DLG, iDLG, and GradInv.


Across FashionMNIST, BettiSafe consistently identifies split locations exhibiting lower reconstruction quality than the heuristic baseline. Similar behaviour is observed on CIFAR-10, where topology-guided selection frequently avoids layers located within empirically identified privacy transition regions. These results indicate that Betti complexity provides a more informative signal than depth alone when selecting privacy-preserving split locations.

The gains are particularly evident in architectures where privacy leakage varies non-monotonically with depth. In such cases, heuristic rules based solely on network position may select layers that remain highly vulnerable to reconstruction attacks, whereas BettiSafe adapts to the underlying representation geometry and favours layers exhibiting reduced topological leakage.

Although the privacy-optimal split varies across architectures and datasets, the overall trend demonstrates that topology-guided selection generalises across diverse settings and consistently identifies competitive privacy-preserving split locations. This suggests that persistent homology can serve as a practical design-time tool for privacy-aware split-layer selection without requiring expensive attack-based evaluation.
\subsection{System-Level Tradeoffs in Split Selection}
\label{subsec:system_tradeoffs}

Beyond privacy leakage, the choice of split layer in SL also affects system-level factors such as communication overhead and inference latency. In practical deployments, split selection must therefore balance privacy protection with computational and communication efficiency. To evaluate these tradeoffs, we analyse the communication cost and latency associated with different split layers for representative architectures and datasets. Communication cost is measured as the dimensionality of smashed activations transmitted between client and server, while latency is measured as the forward-pass computation time up to the split layer.

Table~\ref{tab:system_tradeoffs} summarises the communication size and latency observed across candidate split layers for the FashionMNIST-LeNet configuration. Early split layers produce high-dimensional smashed activations, resulting in larger communication costs but lower client-side computation. Conversely, deeper splits reduce communication requirements while increasing client-side latency. The SSIM values reported in Table~\ref{tab:system_tradeoffs} correspond to the system-level evaluation setting and are therefore not directly comparable with the layer-wise leakage profiling presented in Fig.~\ref{fig:ssim-layerwise}

\begin{table}[htbp]
\centering
\caption{System-level tradeoffs for representative split layers in the FashionMNIST–LeNet configuration. Communication size corresponds to the dimensionality of smashed activations transmitted from client to server. The SSIM column reports the reconstruction quality of feature inversion attacks, where lower values indicate stronger privacy protection.}
\label{tab:system_tradeoffs}
\begin{tabular}{c c c c}
\hline
\textbf{Split Layer} & \textbf{Communication Size} & \textbf{Latency (ms)} & \textbf{SSIM} \\ \hline
3  & 1600 & 0.268 & 0.559 \\
6  & 400  & 0.289 & 0.504 \\
9  & 84   & 0.345 & 0.346 \\
10 & 84   & 0.371 & 0.337 \\ \hline
\end{tabular}
\end{table}

Figure~\ref{fig:privacy_topology_panels} illustrates the relationship between communication cost and inversion leakage across candidate split layers. As expected, early splits exhibit higher leakage due to the richer information content of exchanged representations, whereas deeper splits generally reduce both reconstruction quality and communication cost.

Importantly, BettiSafe generally favours split locations that avoid both highly communication-intensive early layers and privacy-sensitive representations. This behaviour suggests that topology-guided split selection naturally identifies operating regions that provide favourable privacy--efficiency tradeoffs without explicitly optimising communication or latency objectives. Consequently, BettiSafe offers a practical split-selection strategy for collaborative learning deployments where privacy and system performance must be jointly considered.

\subsection{Computational Complexity Analysis}

As discussed in Section~\ref{subsec:betti_leakage}, BettiLeak computation relies on a Vietoris--Rips filtration constructed from $N$ activation vectors and has a worst-case computational complexity of $O(N^2)$ with respect to the number of retained samples.
 In practice, this computation is performed on small calibration mini-batches and bounded by subsampling to at most $N_{\max}$ samples and $D_{\max}$ feature dimensions. Specifically, persistent homology is computed using calibration mini-batches of 64 samples after PCA reduction to 30 dimensions. Under this configuration, Betti profiling requires approximately $9.7,\mathrm{ms}$ per candidate layer on a standard CPU.

Since Betti profiling is performed only once during design-time split selection, the resulting overhead is negligible compared with the cost of model training and deployment. Furthermore, the proposed framework relies exclusively on first-order topological descriptors ($\beta_1$), avoiding higher-dimensional homology computations that are substantially more expensive and often less stable in high-dimensional neural representations.

These results indicate that topology-guided privacy analysis can be incorporated into practical split learning workflows with minimal computational overhead while remaining scalable across architectures and datasets.

\begin{figure*}[!t]
\centering

\begin{minipage}[t]{0.48\textwidth}
\centering

\begin{tikzpicture}
\begin{axis}[
xlabel={Betti Complexity},
ylabel={SSIM},
width=\linewidth,
height=0.65\linewidth,
grid=major,
legend style={at={(0.63,0.99)},anchor=north west,opacity=0.5}
]

\addplot[
only marks,
mark=*,
blue,
mark size=2.5pt
]coordinates {
(0.0031,0.82) (0.0029,0.76) (0.0045,0.34) (0.0061,0.21)
};
\addlegendentry{LeNet}

\addplot[
only marks,
mark=square*,
green!70!black,
mark size=2.5pt
] coordinates {
(0.0038,0.91) (0.0042,0.88) (0.0067,0.39) (0.0094,0.18)
};
\addlegendentry{SimpleCNN}

\addplot[
only marks,
mark=triangle*,
red,
mark size=2.5pt
] coordinates {
(0.0026,0.79) (0.0039,0.41) (0.0078,0.12)
};
\addlegendentry{ResNet-18}

\addplot[
only marks,
mark=diamond*,
orange,
mark size=2.5pt
] coordinates {
(0.004687,0.1254)
(0.014583,0.0532)
(0.010417,0.0442)
(0.007292,0.0582)
};
\addlegendentry{MobileNetV2}

\end{axis}
\end{tikzpicture}

\caption{Relationship between Betti complexity and SSIM-based inversion leakage across architectures on CIFAR-10.}
\label{fig:betti-ssim-crossmodel}

\end{minipage}
\hfill
\begin{minipage}[t]{0.48\textwidth}
\centering

\begin{tikzpicture}

\begin{axis}[
xlabel={Betti-1 Complexity},
ylabel={SSIM Reconstruction Quality},
width=\linewidth,
height=0.65\linewidth,
grid=major,
legend style={at={(0.01,0.4)},anchor=south west,opacity=0.5},
xmin=0,
xmax=0.014,
ymin=0.2,
ymax=0.8,
]

\addplot[
only marks,
mark=*,
blue,
mark size=2.5pt
]
coordinates {
(0.008333,0.5595)
(0.007292,0.5040)
(0.006771,0.3463)
(0.004687,0.3372)
};
\addlegendentry{LeNet}

\addplot[
only marks,
mark=square*,
green!70!black,
mark size=2.5pt
]
coordinates {
(0.0091,0.612)
(0.0082,0.522)
(0.0075,0.401)
(0.0064,0.358)
};
\addlegendentry{SimpleCNN}

\addplot[
only marks,
mark=triangle*,
red,
mark size=2.5pt
]
coordinates {
(0.009375,0.5552)
(0.005729,0.4317)
(0.010417,0.3863)
(0.004687,0.3892)
};
\addlegendentry{ResNet18}

\addplot[
only marks,
mark=diamond*,
orange,
mark size=2.5pt
]
coordinates {
(0.005208,0.3621)
(0.008333,0.2818)
(0.005729,0.2979)
(0.001563,0.3471)
};
\addlegendentry{MobileNetV2}

\end{axis}

\end{tikzpicture}

\caption{Topology--leakage phase diagram illustrating the relationship between Betti-1 complexity and inversion reconstruction quality across architectures on FashionMNIST. Each point represents a candidate split layer. Regions with higher Betti complexity correspond to increased inversion vulnerability, revealing a structural transition in representation geometry associated with privacy leakage.}
\label{fig:phase_diagram_topology_leakage}

\end{minipage}

\end{figure*}







\begin{table}[htbp]
\centering
\caption{Comparison of Betti-guided and heuristic split selection across datasets. $\uparrow$ inversion MSE and $\downarrow$ SSIM indicates stronger privacy. DLG, iDLG, and GradInv metrics are averaged over three implementations to reduce optimisation variance.}
\label{tab:fm-split-compare}
\resizebox{\textwidth}{!}{
\begin{tabular}{lccccccccc}
\hline
\multicolumn{1}{c}{\multirow{2}{*}{Dataset}}                   & \multirow{2}{*}{Model} & \multirow{2}{*}{\begin{tabular}[c]{@{}c@{}}Betti \\ Split\end{tabular}} & \multirow{2}{*}{\begin{tabular}[c]{@{}c@{}}Heuristic\\ Split\end{tabular}} & \multicolumn{2}{c}{MSE}  & \multirow{2}{*}{\begin{tabular}[c]{@{}c@{}}Inv. \\ SSIM\end{tabular}} & \multirow{2}{*}{\begin{tabular}[c]{@{}c@{}}DLG\\MSE\end{tabular}} & \multirow{2}{*}{\begin{tabular}[c]{@{}c@{}}iDLG \\MSE\end{tabular}} & \multirow{2}{*}{\begin{tabular}[c]{@{}c@{}}GradInv\\PSNR\end{tabular}} \\
\multicolumn{1}{c}{} &   &          &             & Betti                 & Heuristic      &   &    &   &   \\ \hline
\multirow{4}{*}{\begin{tabular}[c]{@{}l@{}}FashionMNIST\end{tabular}} & \multicolumn{1}{c}{LeNet}  & \multicolumn{1}{c}{4}         & \multicolumn{1}{c}{6} & \multicolumn{1}{c}{$0.446$} & \multicolumn{1}{c}{$1.06$} & 0.3277 & 0.150655 & 0.158650 & 8.2773 \\
& SimpleCNN & 3        & \begin{tabular}[c]{@{}c@{}}5\end{tabular} & $0.0154$& $0.179$ & 0.5480 & 0.170521 & 0.146742 & 7.5486 \\
& ResNet-18 & 0        & 4           & $0.0321$& $0.267$ & 0.3892 & 0.274937 & 0.282145 & 5.3538\\
& MobileNetV2 & 17 & 17 & $0.0641$ & $0.0597$ & 0.3367 & 0.237963 & 0.235255 & 6.3089 
\\\hline
\multirow{4}{*}{\begin{tabular}[c]{@{}l@{}}CIFAR-10\end{tabular}} & \multicolumn{1}{c}{LeNet}  & \multicolumn{1}{c}{3}         & \multicolumn{1}{c}{5} & \multicolumn{1}{c}{$0.287$} & \multicolumn{1}{c}{$0.694$} & 0.3836 

& 0.182417 & 0.174932 &7.8215\\
& SimpleCNN & 4        & \begin{tabular}[c]{@{}c@{}}6\end{tabular} & $0.0631$& $0.342$ & 0.1321 
& 0.213746 & 0.191583 &7.0214\\
& ResNet-18 & 1        & 4           & $0.0814$& $0.498$ & 0.3864 & 0.298611 & 0.307524 & 5.0412\\
 & MobileNetV2 & 17 & 17 & $0.1123$ & $0.1091$ & 0.0426 & 0.079959 & 0.079381 & 11.1614 \\\hline
\multirow{4}{*}{\begin{tabular}[c]{@{}l@{}}SVHN\end{tabular}} & \multicolumn{1}{c}{LeNet}  & \multicolumn{1}{c}{3}         & \multicolumn{1}{c}{10} & \multicolumn{1}{c}{$0.0035$} & \multicolumn{1}{c}{$0.0459$} & 0.8182 & 0.087067 & 0.100683 & 9.4968 \\
& SimpleCNN & 5        & \begin{tabular}[c]{@{}c@{}}8\end{tabular} & $0.0051$ & $0.0644$ & 0.7412 & 0.084405 & 0.092727 & 9.7825 \\
&  ResNet-18 & 1 & 7 & $0.0053$ & $0.0543$ & 0.7245 & 0.092101 & 0.092163 & 10.0499 \\ 
& MobileNetV2 & 17 & 17 & $0.1107$ & $0.1196$ & 0.1434 & 0.102169 & 0.101100 & 9.8638\\\hline

\end{tabular}
}
\end{table}

\section{Interpreting Privacy Leakage via Representation Topology}
\label{sec:privacy_analysis}

This section provides a representation-level interpretation of the empirical findings presented in Section~\ref{sec:results}. Rather than introducing new experiments, this section examines how the topology of IRs influences privacy leakage in SL and explain the observed behaviour of topology-guided split selection. The discussion focuses on recurring patterns observed across architectures, datasets, and attack settings, providing a structural perspective on why privacy leakage varies across split layers. All observations are grounded in the experimental evidence reported in Section~\ref{sec:results}.
\subsection{Representation-Level Indicators of Privacy Leakage}

The experimental results indicate that privacy leakage in SL is associated with multiple representation-level indicators. As observed throughout Section~\ref{sec:results}, three recurring empirical patterns emerge across architectures, datasets, and split locations:

\begin{itemize}
    \item \textbf{Inversion susceptibility}, reflected by SSIM-based feature inversion quality (Figures~\ref{fig:ssim-layerwise} and~\ref{fig:cifar-ssim-layerwise}).

    \item \textbf{Dataset and architectural sensitivity}, reflected by variations in leakage behaviour across architectures and datasets (Sections~\ref{subsec:cross_model} and~\ref{subsec:cross_dataset}).

    \item \textbf{Topological complexity}, reflected by layer-wise Betti-1 variation in intermediate representations (Figure~\ref{fig:betti-layerwise}).
\end{itemize}

Together, these observations demonstrate that privacy leakage does not evolve monotonically with network depth. Instead, leakage behaviour depends on structural properties of the underlying representations and varies across datasets, architectures, and split locations. This motivates a representation-centric rather than depth-centric perspective on privacy analysis in SL.





\subsection{Topological Complexity as a Proxy for Privacy Risk}

Major observation from Section~\ref{sec:results} is the consistent relationship between Betti-1 complexity and feature inversion leakage. Across multiple architectures and datasets, layers exhibiting elevated Betti-1 values tend to coincide with increased reconstruction quality, whereas layers with low Betti complexity generally correspond to reduced inversion success.

This relationship suggests that Betti-1 captures aspects of representation structure that are relevant to privacy leakage. Since Betti-1 quantifies the presence of persistent loop-like topological features within the activation space, higher values indicate richer geometric organisation of IRs. Such representations retain more structured information about the original inputs and therefore provide a more informative basis for reconstruction-based attacks.

Importantly, the observed relationship extends beyond a single architecture or dataset. As demonstrated in Sections~\ref{subsec:cross_model} and~\ref{subsec:cross_dataset}, the alignment between Betti complexity and inversion vulnerability persists across convolutional, residual, and mobile architectures, as well as across FashionMNIST, CIFAR-10, and SVHN. Although the magnitude of leakage varies between settings, the qualitative trend remains consistent.

These observations do not imply that Betti-1 deterministically predicts attack success. Privacy leakage is influenced by multiple factors, including dataset characteristics, model architecture, and attack methodology. Nevertheless, the empirical evidence suggests that topological complexity serves as a useful representation-level proxy for identifying regions of elevated privacy risk. This provides a structural explanation for the effectiveness of Betti-guided analysis without requiring explicit execution of privacy attacks.





\subsection{Layer-Dependent Privacy Transitions}

Beyond the overall relationship between topological complexity and privacy leakage, the results reveal the presence of distinct privacy transition regions within deep networks. As shown in Section~\ref{subsec:layerwise_privacy}, small changes in split location can lead to disproportionate changes in inversion leakage, producing sharp transitions between privacy-preserving and privacy-sensitive layers.

These transitions are difficult to explain using architectural depth alone. Across both FashionMNIST and CIFAR-10, neighbouring layers often exhibit markedly different reconstruction behaviour despite being separated by only a small number of network operations. This observation suggests that privacy leakage is governed by changes in representation structure rather than by depth itself.

A topological perspective provides a natural explanation for this behaviour. As representations evolve through successive network layers, the geometry of the activation space undergoes periods of relative stability followed by phases of rapid reorganisation. The observed privacy transitions frequently coincide with these structural changes, indicating that inversion vulnerability emerges when representations preserve sufficient geometric organisation to support reconstruction.

This interpretation also explains the limitations of depth-based split selection heuristics. Since privacy-sensitive regions do not occur at fixed depths and vary across architectures and datasets, selecting split layers solely according to network position may overlook critical representational transitions. In contrast, topology-aware analysis identifies changes in representation structure directly, making it better aligned with the empirical behaviour of privacy leakage.

Taken together, these findings suggest that privacy transitions in split learning are fundamentally representation-driven phenomena. Rather than emerging gradually with increasing depth, privacy risk changes most significantly at layers where the underlying geometry of the representation space undergoes substantial structural reorganisation.

\section{Discussion}
\label{sec:discussion}

While Section~\ref{sec:privacy_analysis} provides a representation-level interpretation of the observed privacy behaviours, this section discusses the broader implications of these findings for privacy-aware split learning. We focus on the scope of topology-guided analysis, its practical utility for split-layer selection and training, and the limitations that motivate future research directions.

\subsection{Scope and Limitations of Topological Privacy Analysis}
The empirical findings indicate that the utility of topology-guided privacy analysis depends strongly on the attack model being considered. While topology provides a useful representation-level view of privacy risk, its effectiveness varies across leakage mechanisms and deployment settings. However, the predictive utility of topology is not uniform across attack models. As shown in Section~\ref{subsec:attack_consistency}, gradient-based attacks such as DLG exhibit limited sensitivity to layer-wise topological complexity. This observation suggests that gradient leakage is governed by optimisation dynamics and gradient structure rather than the geometric organisation of smashed activations.

The results further demonstrate that privacy behaviour remains dataset-dependent. While the relationship between network depth and leakage differs substantially between FashionMNIST and CIFAR-10, topology-guided indicators remain informative across both settings. Consequently, Betti complexity appears to capture representation-level properties that generalise beyond specific datasets while remaining sensitive to the underlying privacy regime.

These observations clarify the scope of topology-guided privacy analysis. Rather than serving as a universal predictor of all attack classes, Betti-based descriptors are most effective for identifying privacy risks associated with the geometric structure of intermediate representations.

\subsection{Privacy--Utility Trade-offs via Betti-Regularised Training}

Beyond split-layer selection, the results in Section~\ref{subsec:betti_reg} demonstrate that topological complexity can be directly incorporated into the training objective. Introducing a Betti-based regularisation term increases inversion difficulty while preserving, and in some cases slightly improving, classification accuracy.

This behaviour suggests that excessive topological complexity contributes to the retention of privacy-sensitive structure within smashed activations. By penalising such complexity during training, the resulting representations become less susceptible to reconstruction attacks without significantly degrading predictive performance.

These findings position topology not only as a diagnostic signal for privacy analysis but also as a practical design parameter for privacy-aware representation learning. More broadly, they highlight the potential of topology-guided objectives as an alternative to attack-specific defences that often require substantial modifications to the learning pipeline.

\subsection{Implications for Betti-Guided Split Selection}

A central objective of this work is to investigate whether topological descriptors can support practical split-layer selection. The empirical results indicate that Betti-guided split selection consistently identifies privacy-preserving split locations across multiple architectures and datasets without requiring architecture-specific tuning.

Unlike attack-based evaluation procedures, which require repeated execution of computationally expensive reconstruction attacks, BettiSafe operates directly on intermediate representations. This enables privacy-aware split selection using only forward activations and avoids the need for exhaustive attack simulations during deployment.

The observed generalisation across architectures and datasets further suggests that topology-guided profiling captures structural characteristics of representations that are not tied to a particular model family. Consequently, BettiSafe provides a lightweight and architecture-agnostic mechanism for identifying privacy-sensitive regions and supports privacy-aware deployment across heterogeneous learning environments.
\subsection{Limitations and Future Scope}

While the proposed topology-guided framework demonstrates strong empirical effectiveness, several limitations constrain the scope of the present study and motivate future research.

First, the analysis focuses primarily on feature-space reconstruction attacks, for which topological complexity exhibits strong predictive power. Gradient-based attacks such as DLG show substantially weaker alignment with representation topology, indicating that topology is most informative for privacy threats linked to representation geometry rather than optimisation dynamics. Extending the framework to additional attack families remains an important direction for future work.

Second, although multiple architectures and datasets are considered, the experiments are restricted to image-based learning tasks. Generalisation to higher-resolution imagery, medical imaging, graph data, natural language processing, and tabular learning remains to be investigated.

Third, the current framework relies exclusively on one-dimensional Betti numbers for efficiency and stability. While sufficient to capture the observed privacy transitions, richer descriptors such as persistence entropy, persistence landscapes, or higher-order homological features may provide additional insight.

From a computational perspective, persistent homology is computed on moderate calibration batches to maintain tractability. Developing scalable approximations, streaming formulations, or online topological estimators would improve applicability to large-scale and real-time collaborative learning environments.

Finally, the proposed framework is intentionally empirical and diagnostic in nature. Betti-based analysis provides an attack-agnostic indicator of privacy risk, but does not constitute a formal privacy guarantee. Establishing theoretical links between representation topology, reconstructability, and information leakage remains an open and promising research direction.

Overall, these limitations do not diminish the core contributions of this work. Instead, they highlight opportunities for extending topology-guided privacy analysis toward more comprehensive, scalable, and theoretically grounded split learning systems.

\section{Conclusion}

This paper introduced a topology-guided framework for privacy-aware split learning that leverages persistent Betti complexity to characterise and mitigate privacy leakage in intermediate representations. Through extensive empirical evaluation, we showed that privacy risk in SL is highly non-uniform across layers and exhibits sharp transition regions that are not captured by architectural depth alone. Our results demonstrate that increases in Betti complexity consistently align with the onset of feature inversion vulnerability, with reconstruction fidelity reaching up to 0.98 SSIM at privacy-critical
split points.

Building on this observation, we proposed {BettiSafe}, a topology-guided split selection mechanism that identifies privacy-sensitive layers without requiring attack execution. Across architectures and datasets, Betti-guided splits improved resistance to feature inversion by \textbf{2--5$\times$} compared to heuristic depth-based strategies, while maintaining classification accuracy. Furthermore, Betti-regularised training increased inversion difficulty by nearly \textbf{5$\times$} without degrading model performance, demonstrating a favourable privacy--utility trade-off.

Combined together, these findings establish topological complexity as a robust, representation-level indicator of privacy risk in SL. By shifting split selection from heuristic depth rules to topology-aware structural analysis, the proposed framework provides a practical foundation for privacy-aware SL and demonstrates the broader potential of topological descriptors for guiding collaborative learning system design.

\begin{figure*}[!t]
\centering

\begin{subfigure}[t]{0.32\textwidth}
\centering
\begin{tikzpicture}
\begin{axis}[
xlabel={Split Layer},
ylabel={SSIM},
grid=major,
width=\linewidth,
height=4.5cm,
legend style={
    at={(0.98,0.98)},
    anchor=north east,
    opacity=0.5,
    font=\scriptsize
},
tick label style={font=\scriptsize},
label style={font=\small},
]

\addplot+[mark=*] coordinates {
(3,0.5595)
(6,0.5040)
(9,0.3463)
(10,0.3372)
};
\addlegendentry{LeNet}

\addplot+[mark=square*] coordinates {
(1,0.5552)
(3,0.4317)
(5,0.3863)
(7,0.3892)
};
\addlegendentry{ResNet18}

\addplot+[mark=triangle*] coordinates {
(4,0.5642)
(9,0.1686)
(14,0.0760)
(17,0.1498)
};
\addlegendentry{MobileNetV2}

\addplot+[mark=diamond*] coordinates {
(3,0.612)
(6,0.522)
(9,0.401)
(12,0.358)
};
\addlegendentry{SimpleCNN}

\end{axis}
\end{tikzpicture}

\caption{Reconstruction leakage across architectures on FashionMNIST.}
\label{fig:ssim-architecture-fashion}
\end{subfigure}
\hfill
%
\begin{subfigure}[t]{0.32\textwidth}
\centering
\begin{tikzpicture}
\begin{axis}[
xlabel={Split Layer},
ylabel={Betti-1},
grid=major,
width=\linewidth,
height=4.5cm,
legend style={
    at={(0.98,0.98)},
    anchor=north east,
    opacity=0.5,
    font=\scriptsize
},
tick label style={font=\scriptsize},
label style={font=\small},
]

\addplot+[mark=*] coordinates {
(3,0.008333)
(6,0.007292)
(9,0.006771)
(10,0.004687)
};
\addlegendentry{LeNet}

\addplot+[mark=square*] coordinates {
(1,0.009375)
(3,0.005729)
(5,0.010417)
(7,0.004687)
};
\addlegendentry{ResNet18}

\addplot+[mark=triangle*] coordinates {
(4,0.022917)
(9,0.020573)
(14,0.016667)
(17,0.014583)
};
\addlegendentry{MobileNetV2}

\addplot+[mark=diamond*] coordinates {
(3,0.0091)
(6,0.0082)
(9,0.0075)
(12,0.0064)
};
\addlegendentry{SimpleCNN}

\end{axis}
\end{tikzpicture}

\caption{Topological complexity across architectures on FashionMNIST.}
\label{fig:betti-architecture-fashion}
\end{subfigure}
\hfill
%
\begin{subfigure}[t]{0.32\textwidth}
\centering
\begin{tikzpicture}
\begin{axis}[
xlabel={Split Layer},
ylabel={SSIM},
grid=major,
width=\linewidth,
height=4.5cm,
legend style={
    at={(0.98,0.98)},
    anchor=north east,
    opacity=0.5,
    font=\scriptsize
},
tick label style={font=\scriptsize},
label style={font=\small},
]

\addplot+[mark=*] coordinates {
(3,0.5595)
(6,0.5040)
(9,0.3463)
(10,0.3372)
};
\addlegendentry{FashionMNIST}

\addplot+[mark=square*] coordinates {
(3,0.8208)
(6,0.6014)
(9,0.2198)
(10,0.1581)
};
\addlegendentry{SVHN}

\addplot+[mark=triangle*] coordinates {
(4,0.5642)
(9,0.1686)
(14,0.0760)
(17,0.1498)
};
\addlegendentry{CIFAR-10}

\end{axis}
\end{tikzpicture}

\caption{Dataset-level reconstruction leakage across split layers.}
\label{fig:ssim-dataset-comparison}
\end{subfigure}

\caption{Layer-wise privacy behaviour and topological complexity across architectures and datasets.}
\label{fig:privacy_topology_panels}

\end{figure*}
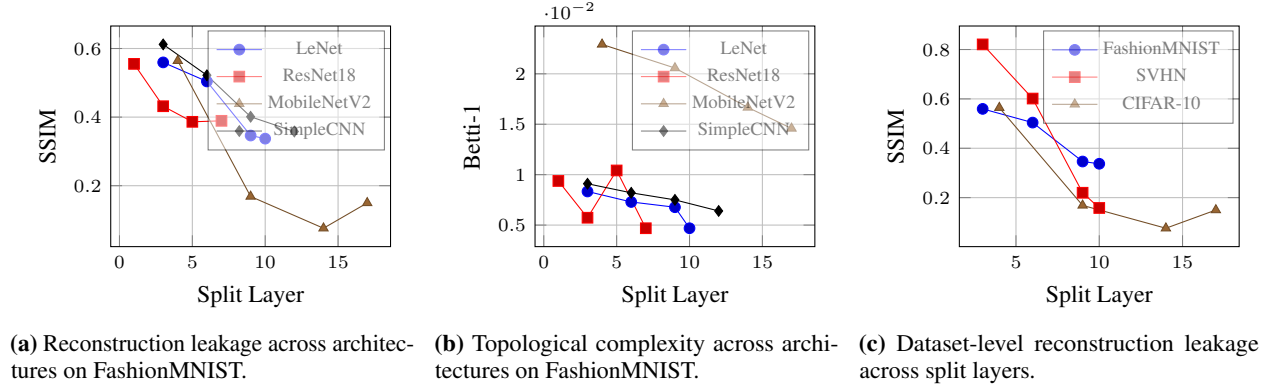

\bibliographystyle{unsrtnat}

\bibliography{cas-refs}





\end{document}